\documentclass[11pt,a4paper]{article}
\usepackage[hyperref]{acl2019}
\usepackage{times}
\usepackage{latexsym}
\usepackage{xcolor}
\usepackage{graphicx}

\usepackage{url}

\aclfinalcopy 

\title{Good Secretaries, Bad Truck Drivers?\\Occupational Gender Stereotypes in Sentiment Analysis}

\author{Jayadev Bhaskaran \\
  ICME, Stanford University, USA \\
  \texttt{jayadevbhaskaran@gmail.com} \\\And
  Isha Bhallamudi \\
  Dept. of Sociology, UC Irvine, USA \\
  \texttt{isha.b@uci.edu} \\}

\date{}

\begin{document}
\maketitle
\begin{abstract}
In this work, we investigate the presence of occupational gender stereotypes in sentiment analysis models. Such a task has implications for reducing implicit biases in these models, which are being applied to an increasingly wide variety of downstream tasks. We release a new gender-balanced dataset\footnote{Link to dataset: \url{https://bit.ly/2HLSKnf}} of $800$ sentences pertaining to specific professions and propose a methodology for using it as a test bench to evaluate sentiment analysis models. We evaluate the presence of occupational gender stereotypes in $3$ different models using our approach, and explore their relationship with societal perceptions of occupations. \end{abstract}

\section{Motivation}
Social Role Theory \citep{eagly1984gender} shows that our ideas about gender are shaped by observing, over time, the roles that men and women occupy in their daily lives. These ideas can crystallize into rigid stereotypes about how men and women ought to behave, and what work they can and cannot do. Gendered stereotypes are powerful precisely for this reason: they define desirable and expected traits, roles and behaviors in people, and go beyond description to prescription. Such biases from the social world, when they map onto machine learning models, serve to reinforce and propagate stereotypes further.

In this paper, we look specifically at occupational gender stereotypes in the context of sentiment analysis. Sentiment analysis is increasingly being applied for recruitment, employee retention and job satisfaction in the corporate world \citep{costa}. Given the prevalence of occupational gender stereotypes, our study primarily deals with the question of whether sentiment analysis models display and propagate these stereotypes. To contextualize and ground our study, we first provide a summary of the relevant sociological literature on occupational gender stereotypes.

\subsection{Background}
Sociological studies as early as 1975 \citep{shinar1975} investigate gender stereotypes of occupations, and rank occupations in terms of how ``masculine", ``feminine" or neutral they are perceived to be. \citet{cejka} successfully predicted the gender distribution of occupations based on beliefs about how specific \emph{gender-stereotypical} attributes (such as ``masculine physical") contribute to occupational success. Such beliefs - that success in a \emph{male} dominated profession, for example, requires \emph{male-specific} traits - directly contribute to sex segregation in occupations. The study also found that high occupational prestige and wages are \emph{strongly correlated} with masculine images. Together, this goes to show that occupational structure is deeply shaped by gender. More recently, \citet{haines} investigate how and whether gender stereotypes have changed between 1983 and 2014, and find conclusive evidence that occupational gender stereotypes have \emph{persisted strongly} through the ages and remain stable. There is ample sociological evidence to show that occupational gender stereotypes have \emph{not} undergone substantial modification since the entry of women into the workplace, and that they remain pervasive and widely held by both men \emph{and} women \citep{glick, haines}.

Since occupational gender stereotypes are shaped by subjective factors and \emph{not} objective reality, they remain resistant to contrary evidence. Theories such as the backlash hypothesis \citep{rudman} further explain their persistence: this theory shows how women in the workplace must disconfirm female stereotypes in order to be perceived as competent leaders, yet traits of ambition and capability in women evoke negative reactions which present a barrier to every level of occupational success. 

The implications of occupational gender stereotypes are profound. Children and adolescents are particularly sensitive to gendered language used to describe occupations and form rigid occupational gender stereotypes based on this \citep{vervecken2013}. In adults, occupational gender stereotypes directly contribute to the existence of unequal compensation and discriminatory hiring. They also lead to self-fulling prophecies: for instance, individuals may not apply to certain jobs in the first place because they think they don't fit the gender stereotype for occupational success in that field \citep{kay}.  

 In the following section, we discuss relevant prior work on gender bias from the NLP literature. In Section ~\ref{methodology} we describe our methodology, dataset, and experiments in greater detail. In Section ~\ref{results}, we present and analyze our results, and finally, Section ~\ref{future} describes possible directions of future work and concludes\footnote{Source code for this paper: \url{github.com/jayadevbhaskaran/gendered-sentiment}}.

\section{Prior Work}
Word embeddings have been the bedrock of neural NLP models ever since the arrival of \texttt{word2vec} \citep{mikolov}, and a variety of topics related to biases with word embeddings have been studied in prior literature. \citet{garg} show the presence of stereotypes in word embeddings through the ages, while \citet{bolukbasi} demonstrate explicit examples of social biases that are introduced into word embeddings trained on a large text corpus. Prior work has also dealt with occupational gender stereotypes in different areas of NLP. \citet{caliskan} formulate a method to test biases (including gender stereotypes) in word embeddings, while \citet{rudinger} investigate such stereotypes in the context of coreference resolution. There have also been efforts to \emph{debias} word embeddings \citep{bolukbasi} and come up with \emph{gender neutral} word embeddings \citep{zhao}. These efforts, however, have attracted criticism suggesting that they do not actually \emph{debias} embeddings but instead \emph{redistribute} the bias across the embedding landscape \citep{gonen}. 

\begin{figure}[ht]
\centering
\includegraphics[scale=.35]{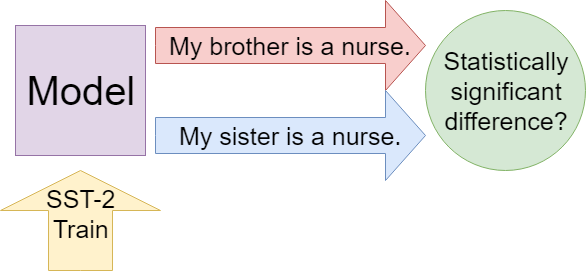}
\caption{Simple diagram of our task definition.}
\label{task_defn}
\end{figure}

Recent trends have been towards replacing fixed word embeddings with large pretrained \emph{contextual} representations as building blocks for NLP tasks. The rise of this paradigm is characterized by the use of language models for pretraining, exemplified by models such as ELMo \citep{elmo}, ULMFit \citep{ulmfit}, GPT \citep{gpt}, and BERT \citep{bert}. 

These models have shown marked improvements over word vector based approaches for a variety of tasks. However, their complexity leads to a tradeoff in terms of interpretability. Recent works have investigated gender biases in such deep contextual representations \citep{may, Basta} as well as their applications to coreference resolution \citep{Zhaonew, webster}; however, no prior work has dealt with such models in the context of occupational gender stereotypes in sentiment analysis. 

\citet{eec} introduce the Equity Evaluation Corpus, a dataset used for measuring racial and gender biases in sentiment analysis-like systems. It was initially used to evaluate systems that predicted emotion and valence of Tweets \citep{SemEval}. We use a similar approach to create a new dataset for measuring gender differences with a specific focus on occupational gender stereotypes. Our approach is \emph{model-independent} and can be used for \emph{any} sentiment analysis system, irrespective of model complexity.

\section{Methodology}\label{methodology}
We create a dataset of $800$ sentences, each with the following structure: \textbf{\texttt{noun} is a/an \texttt{profession}}. Here, \texttt{noun} corresponds to a \emph{male} or \emph{female} noun phrase, such as ``This boy"/``This girl", and \texttt{profession} is one of $20$ different professions. Each sentence is an assertion of fact, and by itself does not seek to exhibit either positive or negative sentiment. Our dataset is balanced across genders and has $20$ noun phrases for each gender, leading to a total of $400$ sentences per gender. 

The rationale behind our selection of the $20$ professions is to include a variety of gender distribution characteristics and occupation types, in correspondence with US Current Population Survey 2018 (CPS) data \citep{cps} and prior literature \citep{haines}. We select $5$ professions that are male-dominated (\emph{truck driver, mechanic, pilot, chef, soldier}) and $5$ that are female-dominated (\emph{teacher, flight attendant, clerk, secretary, nurse}) - with domination meaning greater than $70\%$ share in the job distribution. Next, we add professions that are slightly male-dominated (\emph{scientist, lawyer, doctor}) and slightly female-dominated (\emph{writer, dancer}), with slight domination meaning a $60-65\%$ share in the job distribution. We also add \emph{professor}, which does not have a clear definition as per CPS but has been known to have different gender splits at senior and junior levels. Finally, we include two professions that show an approximately neutral divide (\emph{tailor, gym trainer}) and two which have experienced significant changes in their gender distribution over time (\emph{baker, bartender}), with an increasing female representation in recent times \citep{haines}. As mentioned previously, we also select our set of occupations with an eye towards representing a range of occupation types. 

We evaluate $3$ sentiment analysis models through our experiments. Each model is trained on the Stanford Sentiment Treebank $2$ \texttt{train} dataset \citep{sst}, which contains phrases from movie reviews along with binary ($0/1$) sentiment labels. We then evaluate each model on our new corpus and measure the difference in mean predicted positive class probabilities between sentences with male nouns and those with female nouns. We test $3$ hypotheses (one for each model), with the \emph{null hypotheses} indicating \emph{no difference} in means between sentences with \emph{male} and \emph{female} nouns. Fig. \ref{task_defn} illustrates our experimental setup.

Our evaluation methodology is very similar to that used in \citet{eec}. For each system, we predict the positive class probability for each sentence. We then apply a \emph{paired} t-test (since each pair contains a \emph{male} and \emph{female} version of the same template sentence) to measure if the mean predicted positive class probabilities are different across genders, using a significance level of $0.01$. Since we test \emph{three} hypotheses (one for each system), we apply Bonferroni correction \citep{bonferroni} to the $p$-values that we obtain. In other words, the null hypothesis is rejected only for calculated $p$-values less than $0.01 / 3$. We note that we do not perform any correction to account for the fact that the sentences within each gender are \emph{not} iid, and only vary in the \texttt{noun} and \texttt{profession} words. 

The $3$ models that we evaluate are as follows:
\begin{itemize}
    \item \textbf{M.1}: Bag-of-words + Logistic Regression (baseline): We build a simple bag-of-words model, apply tf-idf weighting, and use logistic regression (implemented using \texttt{scikit-learn} \citep{scikit-learn}) to classify sentiment. This model is a very simple approach that has nevertheless been found to work well in practice for sentiment analysis tasks, and we use it as our baseline model.
    
    \item \textbf{M.2}: BiLSTM: We use a bidirectional LSTM implemented in \texttt{Keras} \citep{keras} to predict sentiment. The words in a sentence are represented by $300$-dimensional GloVe embeddings \citep{glove}. This model is more sophisticated than the baseline and captures some contextual information and long-term dependencies \citep{lstm}. This model also allows us to investigate gender differences that might be introduced through word embeddings, as described in \citet{bolukbasi}.
    
    \item \textbf{M.3}: BERT \citep{bert}: We use a pretrained (uncased) BERT-Base model\footnote{Model source: \url{https://bit.ly/2S8w6Jt}} and finetune it on the SST-2 dataset. This shows near state-of-the-art performance on a wide variety of NLP tasks, including sentiment analysis \citep{bert}.
\end{itemize}

While analysing the results of our experiments, we measure overall predicted mean positive probabilities (across genders) for each of the $20$ professions in our newly created dataset, to identify which professions are rated as \emph{high-sentiment} by these models. This helps us investigate relationships between societal perceptions of occupations and corresponding sentiment predictions from the models.

\begin{table}[ht!]
\begin{center}
\begin{tabular}{|l|r|l|}
\hline \bf Model & \textbf{Dev Acc.} & \bf F - M \\ \hline
\textbf{M.1} (BoW+LogReg) & 0.827 & 0.035** \\
\textbf{M.2} (BiLSTM) & 0.841 & 0.077**  \\
\textbf{M.3} (BERT) & 0.930 & \color{red}-0.040**  \\
\hline
\end{tabular}
\end{center}
\caption{\label{results-table}Results. Dev Acc. represents accuracy on SST-2 \texttt{dev} set. \textbf{F - M} represents difference between means of predicted positive class probabilities for sentences with \emph{female} nouns and sentences with \emph{male} nouns. ** denotes statistical significance with $p < 0.01$ (after applying Bonferroni correction).}
\end{table}

We also examine differences in sentiment among equivalent gender pairs (such as \emph{bachelor} and \emph{spinster}) for the $20$ pairs in our dataset, to investigate differences in predicted sentiment between different sets of male/female noun pairs.    

Finally, we examine differences between \emph{male} and \emph{female} nouns for each individual occupation, to understand which occupations are susceptible to gender stereotyping.

\section{Results/Analysis} \label{results}

The main results of our experiments are shown in Table \ref{results-table}. Our null hypothesis is that the predicted positive probabilities for \emph{female} and \emph{male} sentences have identical means. 
We notice that \textbf{M.1} (Bag-of-words + Logistic Regression) and \textbf{M.2} (BiLSTM) show a statistically significant difference between the two genders, with higher predicted positive class probabilities for sentences with \emph{female} nouns. This effectively represents the biases seen in the SST-2 \texttt{train} dataset. The dataset has $1182$ sentences containing a \emph{male} noun with a mean sentiment of $\mathbf{0.535}$, and $601$ sentences containing a \emph{female} noun with a mean sentiment of $\mathbf{0.599}$. Thus, biases present in training data can get propagated through machine learning models, and our approach can help identify these.

On the contrary, \textbf{M.3} (BERT) shows that sentences with \emph{male} nouns have a statistically significant higher predicted positive class probability than sentences with \emph{female} nouns. One possible reason for this might be biases that propagate from the pretraining phase in BERT. This finding indicates a promising direction of future work: investigating the effects of gender biases in the large pretraining corpus versus those in the smaller fine-tuning corpus (in our case, the SST-2 \texttt{train} dataset). 

\begin{table}[ht!]
\begin{center}
\begin{tabular}{|l|l|}
\hline \bf Model & \textbf{Top 3 professions}  \\ \hline
BoW+LogReg & Secretary, Teacher, Writer \\
BiLSTM &  Dancer, Secretary, Scientist\\
BERT & Scientist, Chef, Dancer \\
\hline
\bf Model & \textbf{Bottom 3 professions}  \\ \hline
BoW+LogReg & Truck Dr., Fl. Att., (many) \\
BiLSTM & Truck Dr., Gym Tr., Nurse \\
BERT & Truck Dr., Clerk, Tailor \\
\hline
\end{tabular}
\end{center}
\caption{\label{professions-table}Top $3$ and bottom $3$ professions per model, based on predicted positive class probability (agnostic of gender). Note: For BoW+LogReg, (many) denotes all the professions that did not appear in the SST-2 \texttt{train} dataset.}
\end{table}

\subsection{Social Stereotypes of Occupations}

We now look at mean distributions of positive class probability (across genders) for each profession, as shown in Table \ref{professions-table}. We notice that \emph{secretary} shows up as a high positive sentiment profession in both \textbf{M.1} and \textbf{M.2}. On further investigation, we notice that this artefact arises because of the 2002 movie \emph{Secretary}, starring Maggie Gyllenhall, that has a number of positive reviews that form a part of the SST-2 \texttt{train} dataset. However, \textbf{M.3} (BERT) seems to be impervious to this, indicating that extensive pretraining could have the potential to remove certain corpus-specific effects that might have lingered in shallower models.  

The profession with the lowest average sentiment score across all $3$ models is \emph{truck driver}; other low scoring professions include \emph{clerk}, \emph{gym trainer} and \emph{flight attendant}. We also note that the highest scoring profession (average sentiment $0.99$) with \textbf{M.3} (BERT) is \emph{scientist} and the lowest (average sentiment $0.34$) is \emph{truck driver}, disturbingly reflective of societal stereotypes about white-collar and blue-collar jobs.

\begin{figure}[ht]
\centering
\includegraphics[scale=.5]{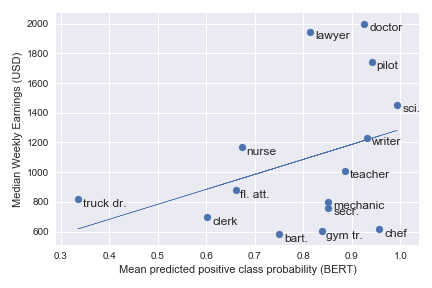}
\caption{Median weekly earnings \citep{cps} vs. mean predicted positive probability using \textbf{M.3} (BERT), per profession.}
\label{plot}
\end{figure}

To explore this further, we look at data from the Current Population Survey of the US Bureau of Labor Statistics \citep{cps}. Fig. \ref{plot} shows the relationship between median weekly earnings (for occupations where data is available) and average positive sentiment predicted by BERT. While there are some outliers, the figure shows a positive correlation between earnings and sentiment, indicating that the model may have incorporated societal perceptions around different occupations. We note that this is only a rough analysis, as not all occupations directly correspond to entries from the survey data.

\subsection{Gendered Stereotypes}
We attempt to analyze differences in gender within occupations by studying the predictions of \textbf{M.3} (BERT), which incorporates the largest amount of external data. First, we analyze differences in mean positive class probability between sentences with male and female nouns for each profession. We notice that \emph{pilot} has the highest \emph{positive} difference between female and male noun sentences (i.e., \emph{female} is \emph{higher}), while \emph{flight attendant} has the \emph{most negative} difference (i.e., \emph{male} is \emph{higher}). This provides an interesting dichotomy: \emph{pilot} is a male-dominated profession, while \emph{flight attendant} is a female-dominated one. 

To test whether these are just artefacts of generic gender bias in the model or specific to occupational gendered stereotypes, we replace \texttt{profession} with ``person" to create $20$ sentence pairs such as ``This man/this woman is a person.", and predict the sentiment for these $20$ pairs. We notice that the difference between \emph{female} and \emph{male} noun sentences for the control experiment is $\mathbf{0.039}$, showing that sentences with \emph{female} nouns in the control group exhibit \emph{higher} positive sentiment that those with \emph{male} nouns. The three occupations with the most negative difference (i.e., \emph{female} sentences have lower positive sentiment) are \emph{flight attendant} ($\mathbf{-0.132}$), \emph{bartender} ($\mathbf{-0.126}$), and \emph{clerk} ($\mathbf{-0.116}$). Of these, \emph{flight attendant} ($72\%$) and \emph{clerk} ($86\%$) are female-dominated professions \citep{cps}, while \emph{bartender} ($55\%$) is a profession that has been shifting from male to female-dominated in recent times \citep{haines}.

Finally, we study differences between corresponding pairs of female and male nouns, using predictions from \textbf{M.3} (BERT). Out of the $20$ pairs in our dataset, the pair with the greatest difference in mean positive class probability is \emph{spinster} and \emph{bachelor}, with $\mathbf{spinster - bachelor = -0.404}$ ($p < 0.01$). This reflects societal perceptions of \emph{spinster} as someone who is characterized as alone, lonely and resembling an ``old maid", versus \emph{bachelor} as someone who might be young, carefree and fun-loving \citep{neiuwets}. This is an example of \emph{semantic pejoration} seen in society, where the female form of the noun (i.e., \emph{spinster}) gradually acquires a negative connotation. Notably, this pejorative behavior may have also leaked into the model, reflecting societal gender stereotypes.

\section{Conclusion/Future Work} \label{future}
In this paper, we introduce a new dataset that can be used to test the presence of occupational gender stereotypes in \emph{any} sentiment analysis model. We then train $3$ sentiment analysis models and evaluate them using our dataset. Following that, we analyze our results, exploring social stereotypes of occupations as well as gendered stereotypes. We find that all $3$ models that we study exhibit differences in mean predicted positive class probability between genders, though the directions vary. We also see that simpler models may be more susceptible to biases seen in the training dataset, while deep contextual models may exhibit biases potentially introduced during pretraining.

One promising avenue for future work is to explore occupational stereotypes in deep contextual models by analyzing their training corpora. This could also help identify techniques to mitigate biases in such models, since they could be relatively impervious to biases introduced by fine-tuning (especially on smaller datasets). 

From a sociological perspective, we plan to investigate occupational gender stereotypes in downstream applications such as automated resume screening. Such a task assumes greater importance with the increased use of these systems in today`s world. There is prior work on ethnic bias in such tools \citep{derous}, and we believe that there is significant value in exploring and characterizing gender biases in these systems.

\section*{Acknowledgements}
We thank Johan Ugander for helping motivate the initial phases of this work. We also thank the anonymous reviewers for their thoughtful feedback and suggestions.

\bibliography{acl2019}
\bibliographystyle{acl_natbib}

\end{document}